\documentclass{article}

\usepackage{graphicx} % more modern
\usepackage{subfigure}

\usepackage{algorithm}
\usepackage{algorithmic}
\usepackage{appendix}

\usepackage{hyperref}

\input{macros}

\def\A{{\bf A}}
\def\a{{\bf a}}
\def\B{{\bf B}}

\def\C{{\bf C}}

\def\D{{\bf D}}

\def\K{{\bf K}}
\def\H{{\bf H}}

\def\I{{\bf I}}
\def\R{{\bf R}}
\def\X{{\bf X}}
\def\Y{{\bf Y}}

\def\S{{\bf S}}
\def\x{{\bf x}}
\def\y{{\bf y}}
\def\z{{\bf z}}
\def\Z{{\bf Z}}
\def\M{{\bf M}}

\def\0{{\bf 0}}
\def\1{{\bf 1}}

\def\RB{{\mathbb R}}

\def\eg{{\emph{e.g. }}}
\def\ie{{\emph{i.e. }}}

\def\etal{{\em et al.\/}\,}

\graphicspath{{figure/}}

\title{Multi-Level Feature Descriptor for Robust Texture Classification via Locality-Constrained Collaborative Strategy}

\author{
Shu Kong and Donghui Wang\\
\texttt{\{aimerykong, dhwang\}@zju.edu.cn}\\
College of Computer Science and Technology, Zhejiang University\\
Hangzhou, China
}

\date{\today}

\begin{document}
\maketitle

\begin{abstract}
This paper introduces a simple but highly efficient ensemble for robust texture classification, which can effectively deal with translation, scale and changes of significant viewpoint problems.
The proposed method first inherits the spirit of spatial pyramid matching model (SPM), which is popular for encoding spatial distribution of local features, but in a flexible way, partitioning the original image into different levels and incorporating different overlapping patterns of each level.
This flexible setup helps capture the informative features and produces sufficient local feature codes by some well-chosen aggregation statistics or pooling operations within each partitioned region, even when only a few sample images are available for training.
Then each texture image is represented by several orderless feature codes and thereby all the training data form a reliable feature pond.
Finally, to take full advantage of this feature pond, we develop a collaborative representation-based strategy with locality constraint (LC-CRC) for the final classification, and experimental results on three well-known public texture datasets demonstrate the proposed approach is very competitive and even outperforms several state-of-the-art methods. Particularly, when only a few samples of each category are available for training, our approach still achieves very high classification performance. % \eg $90.0\%$ average accuracy for Brodatz dataset when only one sample image is used for training, significantly higher than any other methods.
\end{abstract}

\section{Introduction}
\label{intro}

\begin{figure}[t]
\begin{center}
   \includegraphics[width=0.7\linewidth]{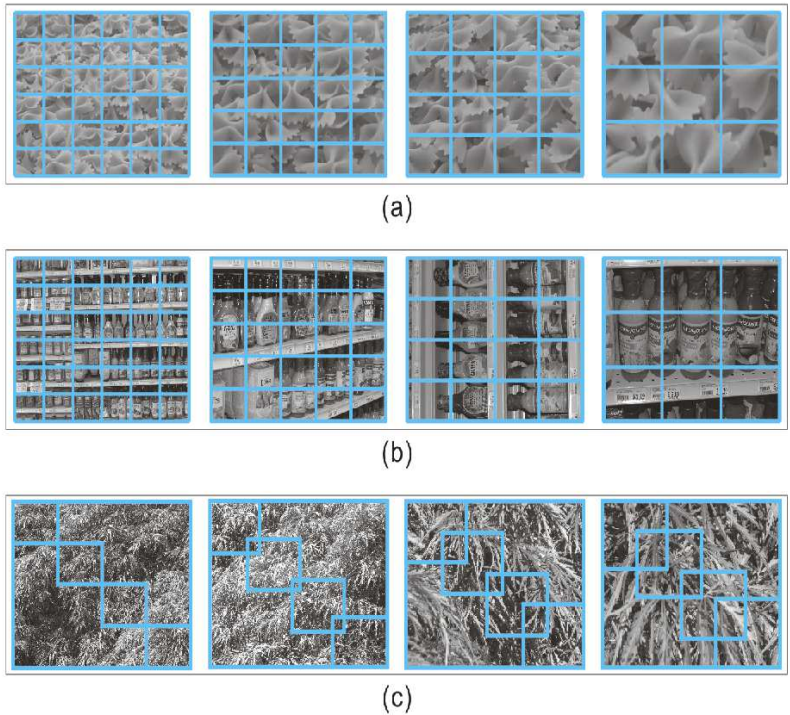}
\end{center}
   \caption{Several samples of three categories from UMD texture database~\cite{Xu_IJCV2009}.
   %Three rows display our partition strategy and different overlapping patterns of a single one partition level.
   From (a) and (b), we can see the statistic information within of the regions from various partition levels can capture  multiple-scale feature information. As a result, local scale and translation differences can be effectively alleviated.
   (c) presents different overlapping patterns in $4 \times 4$ partition. Here only the diagonal regions are plotted for better illustration. This overlapping partition strategy helps us capture reliable and redundant information of the textures.
   %Allowing various overlapping patterns helps better capture deep informative features when only a few samples of each class are available.
   }
\label{fig:sampleShow}
\end{figure}

Texture is widely considered as a fundamental ingredient of the structure of natural images, and texture classification is an important problem in computer vision with many applications. Yet despite almost 50 years of research and development, designing a high-accuracy and robust texture classification system for real-world applications remains a challenge for at least three reasons: the wide range of various natural texture types; the presence of large intra-class variations in texture images, such as rotation, scale, viewpoint, and even non-rigid surface deformation, caused by arbitrary viewing and illumination conditions; and the demands of low computational complexity and a desire to limit algorithm tuning~\cite{WaveletPyramid_CVPR2010}.

There are four basic elements that constitute a reliable texture classification system, as Liu \etal point out in~\cite{SortedRandomProjection_Texture_ICCV2011}: (1) local texture descriptors, (2) non-local statistical descriptors, (3) the design of a distance/similarity measure, and (4) the choice of classifier. Thanks to the emergence of \emph{Bag-of-Feature} words (BoF) model, which treats an image as a collection of unordered appearance descriptors extracted from local patches, quantizes them into discrete "visual words" and then computes a compact histogram representation for semantic image classification. As a result, recent interest for texture classification tends to represent a texture non-locally by the distribution of local textons~\cite{Lazebnik_PAMI2005, LocalFeatures_IJCV2007, basicFeatureforTextureClassification_IJCV2010, WaveletPyramid_CVPR2010}, and achieves state-of-the-art performance.

As an extension of BoF, \emph{spatial pyramid matching} model (SPM)~\cite{Lazebnik_CVPR06} has emerged as a popular framework to represent an image by extracting image descriptors such as SIFT~\cite{Lowe_IJCV2004} or HOG~\cite{HOG_cvpr2005} on a dense grid, encoding them over a learned dictionary, and then summarizing the distribution of the codes in the cells of a spatial pyramid by some well-chosen aggregation statistics, or pooling operation. SPM paradigm has made a remarkable success on a range of image classification benchmarks, and becomes a major component of the state-of-the-art systems~\cite{SPM_Yang_cvpr2009, Boureau_locals_ICCV2011, discriminativeSpatialPyramid_cvpr2011}. Inspired by SPM, we introduce a similar framework to SPM to partition an image into increasingly fine segments, but in a more flexible way, exploiting multi-level partitions with various overlapping patterns and thereby forming redundant local texture feature codes for each regions by a pooling operation. In this way, our method produces a reliable feature pond containing these informative feature codes, even when only a few samples of each class are available for training.

To take full advantage of the feature pond, we develop a simple but effective and efficient mechanism for the final classification, called \emph{collaborative representation-based classification with locality constraint}, LC-CRC for short, which is similar in appearance to \emph{sparse representation-based classification} (\emph{SRC})~\cite{SRC_pami2009}, but essentially differs in two ways: (1) $\ell_{2}$-norm regularization is adopted in the least square fitting problem rather than $\ell_{1}$-norm penalty, and (2) locality constraint is employed to speed up classification process.

We summarize the advantages of our texture classification system below:
\begin{itemize}
  \item Different from many state-of-the-art texture classification methods which combine several types of descriptors, our approach uses only a single type of feature descriptor, \ie SIFT descriptor~\cite{Lowe_IJCV2004}. Thus, our method is much simple but still much capable of discriminating textures demonstrated in the experiment.
  \item Benefiting from the flexible partition strategy, the proposed method can produce redundant feature codes to form a reliable feature pond, even though only a few samples of each category are available for training.
  \item Instead of the widely-used SVMs, a simple but effective classification mechanism, LC-CRC , is developed in our method. It facilitates overall translation, scale, and viewpoint invariance in classification. And experimental results demonstrate the suggested LC-CRC is very effective and efficient.
\end{itemize}

The rest of this paper is organized as follows. Section~\ref{sec:relatedWork} gives a very brief review on the related work. Details of our framework are elaborated in Section~\ref{sec:GeneralFramework}. We show experimental results in Section~\ref{sec:exp} and conclude with some discussion in Section~\ref{sec:conclusion} before closing.

%------------------------------------------------------------------------
\section{Related Work}
\label{sec:relatedWork}

\begin{figure*}
\begin{center}
\includegraphics[width=1\linewidth]{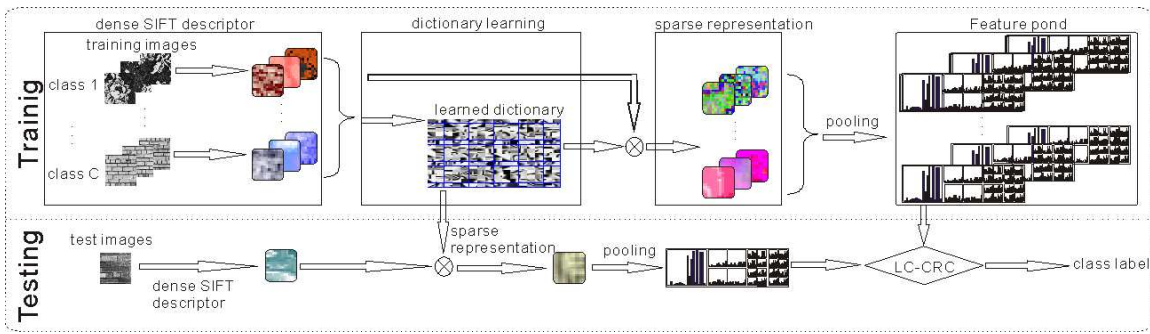}
\end{center}
   \caption{The flowchart of our proposed robust texture classification approach. }
\label{fig:flow}
\end{figure*}

We review closely related work on spatial pyramid matching (SPM), sparse representation-based classification paradigm (SRC), and local coordinate coding (LCC).

SPM framework has made a remarkable success on a range of image classification benchmarks, and remains a major component of the state-of-the-art systems~\cite{Grauman_ICCV05, Lazebnik_CVPR06, Boureau_MidLevelFeature_CVPR2010, Boureau_locals_ICCV2011, discriminativeSpatialPyramid_cvpr2011}.
In SPM, spatial order of local descriptors is seriously considered in image classification tasks, however, it is of little importance and captures no essential features in texture classification, because texture can be coined as "a subtle balance between repetition and innovation"~\cite{WhatIsTexture}, or approximative reduplication. Despite of this, SPM indeed inspires us to borrow the idea of multi-level scheme (pyramid partition of the image) to design a local invariant framework for texture classification. This is one focus of our work --- adopting a more flexible partition configuration with multiple overlapping patterns such as  $\{3, 4, 5\}$\footnote{ It means an image is partitioned into $3\times3$ grid cells in the first level, $4\times4$ and $5\times 5$ for the second and third level, respectively. Altogether $3\times 3 + 4\times 4 + 5 \times 5 = 50$ sub-images or regions or grid cells are formed.}, as Figure~\ref{fig:sampleShow} shows, in place of the common fashion of SPM, \eg $\{1, 2, 4\}$ in \cite{SPM_Yang_cvpr2009} and $\{1, 2, 4, 8\}$ in \cite{discriminativeSpatialPyramid_cvpr2011}, .

Despite the widely-used SVM classifier, \emph{sparse representation-based classification} (SRC)~\cite{SRC_pami2009} achieves a great success in face recognition task, and it boosts the research of sparsity based pattern classification. SRC solves a $\ell_1$-norm penalized least square problem and identifies the class label individually, by checking class by class. However, $\ell_1$-minimization makes SRC very computationally expensive, and recent research shows it lacks stability in face recognition~\cite{FaceRec_CSscheme_CVPR2011, FaceRec_SRorCollaborate_CVPR2011}. Furthermore, Zhang \etal point out the truth that SRC can produce interesting outcome is the contribution of collaborative representation~\cite{FaceRec_SRorCollaborate_CVPR2011}, and they propose a new $\ell_2$ regularized classification algorithm called \emph{collaborative representation-based classification with regularized least square} (CRC-RLS), which adopts $\ell_2$-norm penalty rather than $\ell_1$ regularizer in SRC. This modification leads to the simple ridge regression.
However, when the number of collaborative data (training data) grows, calculating the coefficients by ridge regression becomes more computationally expensive, because a larger matrix inverse operation is involved.

Luckily, in a parallel process, focusing on high-dimensional sparse coding problem, Yu \etal empirically observe that sparse coding results tend to be local -- nonzero coefficients are often assigned to bases nearby the encoded data. And they theoretically point out that under certain assumptions locality is more essential than sparsity~\cite{NonlinearLLC_NIPS2009}. To make full use of this local relation, they suggest a modification to sparse coding, called \emph{local coordinate coding} (LCC), and based on their work, Wang \etal propose a practical method called \emph{locality-constraint linear coding} (LLC) to fast implement LCC, and approximate LLC by utilizing \emph{K-nearest neighbors} (KNN) ahead of time~\cite{LLC_Wang_cvpr2010}.

Combining the CRC-RLS and the idea of LLC, we develop a new classification mechanism, which adopts collaborative representation-based recipe regularized by $\ell_2$ penalty and employs KNN search beforehand. Therefore, our classification approach is more stable to outliers as stated in~\cite{FaceRec_CSscheme_CVPR2011}, and much more efficient because only small-size ridge regression is involved even when the number of training samples is large.

%------------------------------------------------------------------------
\section{The proposed Texture Classification Framework}
\label{sec:GeneralFramework}
Focusing on the four basic elements of a reliable texture classification system, in this section, we introduce our proposed framework in detail: local texture descriptors, overall texture image representation, measurement and classification mechanism. The overall flowchart of our approach is displayed by Figure~\ref{fig:flow}. Notations used in this paper are embedded in Subsection~\ref{ssec:descriptor}.

\subsection{Local Texture Descriptor}
\label{ssec:descriptor}

In our work, we use a single type of feature descriptor, the popular SIFT descriptor~\cite{Lowe_IJCV2004}, which is extracted on a dense grid rather than at interest points and has been shown to yield superior classification performance in~\cite{SPM_Yang_cvpr2009, LLC_Wang_cvpr2010, Boureau_MidLevelFeature_CVPR2010, Boureau_locals_ICCV2011}.
Suppose there are $T$ images from $C$ classes and ${\cal I}_{c}$ denotes the index of $c^{th}$ class, and let $t^{th}$ image be represented by a set of dense SIFT descriptors $\x_{i}^{(t)} \in \RB^{d}$ ($d=128$ for SIFT descriptor) at $N$ locations identified with their indices $i=1,\dots,N$. $M$ regions of interest are defined on the image with ${\cal N}_{m}$ denoting the set of locations/indices within region $m$, and $m \in {\cal L}_{l}$ means $m^{th}$ region belongs to $l^{th}$ level, \ie ${\cal L}_{l}$ indexes the regions in $l^{th}$ level. Then we use all the dense SIFT descriptors to train a dictionary $\D \in \RB^{d \times D}$, and employ the learned dictionary back to represent the dense SIFT descriptors into a sparse code vector, as the formulation below:
\begin{equation}
\label{equa:DL}
\begin{split}
( \a_{i}^{(t)}, \D) &= \arg\min\limits_{\a_{i}^{(t)}, \D} \sum\limits_{t=1}^{T}\sum\limits_{i=1}^{N}
\{\Vert \x_{i}^{(t)} - \D \a_{i}^{(t)} \Vert_{2}^{2} + \lambda \Vert \a_{i}^{(t)} \Vert_{1}\}\\
&\text{s.t. } {\bf d}^{T}_{i}{\bf d}_{i} \le 1 \text{ for $i=1,\dots,D$}.
\end{split}
\end{equation}
where $\a_{i}^{(t)} \in \RB^{D}$ is the corresponding sparse code vector.

Each element $a_{k}$ of the code vector $\a$ indicates the local descriptor's response to the $k^{th}$ visual word in the dictionary $\D$. We align all the SIFT descriptors belonging to region $m$ as a matrix $\X \in \RB^{d \times |{\cal N}_{m}|}$, then the corresponding code matrix $\A \in \RB^{D \times |{\cal N}_{m}|}$ is obtained. Here we aggregate the local descriptors' responses across all the $|{\cal N}_{m}|$ locations of this region into an $|{\cal N}_{m}|$-dimensional response vector ${\bf a}^{T}_{k}$ (the $k^{th}$ row of $\A$), in which each elements $a^{T}_{k,m}$ of ${\bf a}^{T}_{k}$ represents the response of the local descriptor $\x_{m}$ at the $m^{th}$ location to the $k^{th}$ visual word. After obtaining all the feature descriptors $\A$ within a region, we can use a pooling operation to pool these feature descriptors  into a single vector $\y$ of fixed dimension, described in Subsection~\ref{sssec:featurePooling}. Before feature pooling, we first address the relevant partition issues.

\subsubsection{Partition Issues}
Different from classical and commonly used SPM scheme, which is three or four level pyramid comprising pooling regions of $\{1 \times 1, 2 \times 2, 4 \times 4\}$ or $\{1 \times 1, 2 \times 2, 4 \times 4, 8 \times 8 \}$~\cite{discriminativeSpatialPyramid_cvpr2011}, we adopt a more flexible partition strategy and divide the original image into finer regions, \eg $\{3\times 3, 4\times4, 5 \times5 \}$ as Figure~\ref{fig:sampleShow} shows.

Merely relying on this flexible partition fashion, through our observation, the proposed method can indeed capture local features in different scales and is resilient to local variance, such as translation, illumination and scale. But we go beyond by permitting different overlapping patterns at the same level of pyramid. Various overlapping patterns within a single level produce more regions when adopting the same partition pattern, \eg a single level of $4\times4$ partition with 4 overlapping patterns will lead to $4\times4\times4 = 64$ regions, as displayed by row~(c) in Figure~\ref{fig:sampleShow}, and accordingly $64$ feature codes will be formed. More overlapping choices can produce more local texture features on multiple scales, and therefore these redundant local texture features can effectively alleviate the classification challenge caused by local variance.

This way of partition with multiple overlapping patterns prevents the statistical information or pooled feature codes of local regions from becoming too rigid or too flappy for texture discrimination, and
in conjunction with our proposed classification mechanism described in Subsection~\ref{ssec:Classification}, it will lead to state-of-the-art performance of texture classification in the experiments.

\subsubsection{Feature Pooling}
\label{sssec:featurePooling}
Feature pooling is essentially to map the response vectors within each region into a statistic value $f(\a^{T}_{k})$ via some spatial pooling operation $f$. Among various pooling methods, such as average pooling, max pooling and some other pooling methods transiting from average to max~\cite{Boureau_ICML2010}, max pooling is inspired by the mechanism of the complex cells in the primary visual cortex and has been shown a powerful operation empirically and theoretically~\cite{SPM_Yang_cvpr2009, Boureau_ICML2010, Boureau_locals_ICCV2011, Boureau_MidLevelFeature_CVPR2010}. In this paper, we also adopt max pooling for its translation-invariance in different level of partitions~\cite{SupTranInvariantSC_CVPR2010}.

After obtaining code matrix $\A$ of region $m$, we can pool the code vectors into one feature vector $\y_{m} \in \RB^{D}$ to represent region $m$:
\begin{equation}
\label{equa:maxPooling2}
\begin{split}
\y_m({\A})& = [f(\a^{T}_{1}), \dots, f(\a^{T}_{k}), \dots, f(\a^{T}_{K})]^{T} \\
& = [ \max\limits_{i \in {\cal N}_{m}} a_{1,i},  \dots, \max\limits_{i \in {\cal N}_{m}} a_{k,i},\dots, \max\limits_{i \in {\cal N}_{m}} a_{K,i}]^{T}\\
\end{split}
\end{equation}
Actually, no matter how the size of different regions differs, the pooled feature code is of the same dimension and well summarize the distribution of the SIFT feature descriptors in each region. This property enables us to adopt the flexible partition way and various overlapping patterns within the same level of partition, thereby producing redundant local texture features.

\subsection{Texture Image Descriptor}
\label{ssec:ImageDescriptor}
As described in the previous subsection, we store all the pooled feature codes of one image to form a matrix $\Y = [ \y_{1}, \dots, \y_{m}, \dots, \y_{M}]$ as the new texture image representation.
That is to say, regardless of region size and overlapping patterns, all the pooled feature vectors of regions are stored in an orderless way.
This orderless storage, in conjunction with max pooling, enjoys translation and scale invariance.
From Figure~\ref{fig:sampleShow}, it is not difficult to see the samples from the same class can represent one another by the statistic information that max pooling accumulates, which is local translation and scale invariant, therefore overall invariance property can be attained. We will see the benefit of this orderless storage from experiment in Section~\ref{sec:exp}.

\subsection{Measure and Classification}
\label{ssec:Classification}
Actually, all the pooled feature vectors from regions of various levels of training images can be seen as redundant feature bases, or a feature pond, which can effectively represent pooled feature codes of a new image, and in this way, scale and translation invariance can be achieved. To fully take advantage of the benefit of orderless feature vector storage, we utilize a regularized least square (RLS) framework for the final classification. It is similar in appearance to sparse representation-based classification (SRC)~\cite{SRC_pami2009}, but essentially different.

In SRC, a vectorized test image $\z$ is coded collaboratively over the dictionary of all $T$ training samples $\Y=[\y_{1}, \dots, \y_{t}, \dots, \y_{T}]$ under $\ell_{1}$-norm sparsity constraint, where $\y_{t}$ is $t^{th}$ vectorized training sample. For simplicity, SRC first calculate sparse coefficients by the formulation:
\begin{equation}
\label{equa:SRC}
\begin{split}
\a = \arg\min\limits_{\a} \Vert \z - \Y \a\Vert_{2}^{2} + \lambda \Vert \a \Vert_{1}
\end{split}
\end{equation}
Then, SRC classifies test image $\z$ individually to determine which class $\z$ should belong to. In other words, it calculates reconstruction error $r_{c} =  \Vert \z - \Y_{c}\a_{c} \Vert_{2}$ for all the $C$ classes, where $\Y_{c}$ is formed by the columns indexed by ${\cal I}_{c}$ and $\a_{c}$ is formed in the similar way. Finally it selects $\hat{c} = \arg\min\limits_{c}r_{c}$ as the predicted label.

Although SRC has shown interesting results in face recognition and has been widely studied in the community, researchers recently have found that, in SRC, $\ell_{1}$-norm penalty in Equation~\ref{equa:SRC} actually makes the classification framework unstable~\cite{FaceRec_SRorCollaborate_CVPR2011, FaceRec_CSscheme_CVPR2011}, as well as computationally very expensive. Zhang \etal point out the truth that SRC improves face recognition accuracy is the use of collaborative representation, but not $\ell_{1}$ sparsity~\cite{FaceRec_SRorCollaborate_CVPR2011}. And they propose a collaborative representation-based classification framework with regularized least square (CRC-RLS) by solving a ridge regression formulation:
\begin{equation}
\label{equa:CRC_RLS}
\begin{split}
\a = \arg\min\limits_{\a} \Vert \z - \Y \a\Vert_{2}^{2} + \lambda \Vert \a \Vert_{2}^{2},
\end{split}
\end{equation}
Following the rest part of SRC, CRC-RLS achieves very competitive classification results but with significantly less complexity than SRC.

However, when the number $T$ of training samples grows, calculating the coefficients by ridge regression $\a = (\Y^{T}\Y + \lambda\I)^{-1} \Y^{T}\z$ becomes more computationally expensive,
because inverse operation on a larger matrix of size $T\times T$ is involved.
To circumvent this problem, we borrow the idea of LLC~\cite{LLC_Wang_cvpr2010} described in Section~\ref{sec:relatedWork} by applying KNN search among the feature pond before solving the ridge regression ---  choosing $K$ nearest neighbors to form $\Y_{(K)} \in \RB^{D \times K}$ with indices ${\cal H}_{(K)}$, and representing the testing image by solving a much lower-complexity ridge regression: $ \hat{\a} = (\Y_{(K)}^{T}\Y_{(K)} + \lambda\I)^{-1} \Y_{(K)}^{T}\z$. After this, an overall coefficient vector $\a\in \RB^{T}$ is formed by embedding the elements of $\hat{\a} \in \RB^{K}$ in ${\cal H}_{(K)}$ locations of ${\a}$ and zeros elsewhere. The final classification follows SRC, and Algorithm~\ref{alg:LC_CRC} shows the whole classification algorithm\footnote{ Because of one texture image is represented by a descriptor matrix as Subsection~\ref{ssec:ImageDescriptor} introduces, in the algorithm, each column of descriptor matrix should be treated individually, and the final reconstruction error is to accumulate over $L$ columns (each column denote a pooled feature code of a specific region) --- the summation of the smallest error of each level. Empirically, we find the using of smallest error for classification brings out better performance than that of the reconstructive errors of all the codes.}.

\begin{algorithm}[t]%H]
\small
\caption{Algorithm of LC-CRC}
\label{alg:LC_CRC}
\emph{Input}: feature descriptor matrix of testing image $\Z = [\z_{1},\dots,\z_{M}]$ and feature pond formed by all the training samples $\Y=[\Y_{1},\dots,\Y_{t},\dots,\Y_{T}]$, where $\Y_{t} = [\y_{1},\dots,\y_{M}]$, parameter $K$ for KNN search and $\lambda$ for balancing $\ell_{2}$-norm penalty and least square fitting.

\emph{Output}: predicted label of the test image.

\begin{algorithmic}[1]
\STATE  Normalize the columns of $\Z$ and $\Y$ to have unit $\ell_{2}$-norm length;
\FOR{$m=1,2,\dots,M$}
    \STATE Use KNN within feature pond $\Y$, selecting $K$ neighbors of $\z_{m}$ to form matrix $\Y_{(K)} \in \RB^{D\times K}$ with $K$ indices ${\cal H}_{m}$;
    \STATE Code $\z_{m}$ over $\Y_{(K)}$ by
    \begin{displaymath}
            \hat{\a}^{m} = (\Y_{(K)}^{T}\Y_{(K)}+ \lambda\I)^{-1}\Y_{(K)}^{T}\z_{m};
        \nonumber
    \end{displaymath}
    \STATE  Form $MT$-dimensional vector $\a^{m}$ where elements of ${\cal H}_{m}$ locations are embedded with $\hat{\a}^{m}$ and zeros elsewhere;
\ENDFOR
\STATE Compute the reconstruction error for each class:
    \begin{displaymath}
        \begin{split}
            r_{c} = \sum\limits_{l=1}^{L} \{ \min\limits_{m\in{\cal L}_{l}}\Vert \y_{m} - \Y_{c}{\hat{\a}_{c}^{m}}\Vert_{2} \};
        \end{split}
    \end{displaymath}
\STATE Output the identity of test image $\Y$ as:
\begin{displaymath}
    \begin{split}
        identity(\Y) = \arg\min\limits_{c}(r_{c}).
    \end{split}
\end{displaymath}
\end{algorithmic}
\end{algorithm}

%------------------------------------------------------------------------
\section{Experiment}
\label{sec:exp}
We evaluate the performance of the proposed texture classification framework on three public datasets: Brodatz dataset~\cite{Brodatz1966}, KTH-TIPS dataset~\cite{KTH_TIPS_ECCV2004} and UMD texture database~\cite{Xu_IJCV2009}.

The Brodatz dataset is a well-known benchmark database for evaluating texture recognition algorithms. It contains 111 different texture classes. For each class, it is represented by only one sample, which is then divided into 9 sub-images non-overlappingly to form the database. Thus, there are 999 images altogether with resolution of 215x215. Although this dataset lacks interclass variations, Lazebnik \etal point out that this database is a challenging platform for testing the discriminative power of texture descriptors, thanks to its variety of scales and geometric patterns~\cite{Lazebnik_PAMI2005}.
The KTH-TIPS textures dataset contains ten texture classes. Images are captured at nine scales spanning two octaves (relative scale changes from 0.5 to 2), viewed under three different illumination directions and three different poses, thus giving a total of 9 images per scale, and 81 images per material class. Some sample images are shown in Figure~\ref{fig:similarPattern_KTHTIPS}, and we can easily see the scaling and illumination changes increase the intra-class variability and makes this database especially difficult for classification task.
UMD texture database is composed of 25 different texture classes, 40 samples for each, and all images are grayscale of 1280x960 pixels (1000 samples altogether). The textures are acquired under strong viewpoint and scale changes, arbitrary rotations, and significant contrast differences, even including textures with nonrigid deformation. Figure~\ref{fig:sampleShow} displays some sample images from this database.

%------------------------------------------------------------------------
\subsection{Configurations and Implementation}

\textbf{Dictionary learning and sparse coding. }
As a dictionary learning problem, Equation~\ref{equa:DL} is convex in $\a_{i}^{(t)}$ with fixed $\D$ and vice versa, but not convex simultaneously for both of them. The conventional way for such problem is to solve it iteratively by alternately optimizing over $\D$ or $\a_{i}^{(t)}$'s while fixing the other. Fixing $\D$, the optimization can be solved by optimizing over each coefficient $\a_{i}^{(t)}$ individually:
\begin{equation}
\begin{split}
\min\limits_{\a_{i}^{(t)}} \Vert \x_{i}^{(t)} - \D \a_{i}^{(t)} \Vert_{2}^{2} + \lambda \Vert \a_{i}^{(t)} \Vert_{1}
\end{split}
\nonumber
\end{equation}
This is essentially a linear regression problem with $\ell_{1}$-norm regularization on the coefficients, \ie Lasso in the Statistical literature. In our work, we solve this optimization by a very efficient algorithm called \emph{feature-sign search}~\cite{efficientsparsecoding_nips}. Fixing all the $\a_{i}^{(t)}$'s, the problem is reduced to a least square problem with quadratic constraints:
\begin{equation}
\begin{split}
\min\limits_{\D} & \Vert \X - \D \A \Vert_{F}^{2},
\text{ s.t. $\Vert {\bf d}_{i}\Vert_{2} \le 1$ for $i=1,\dots,D$.}
\end{split}
\nonumber
\end{equation}
The optimization can be done efficiently by the Lagrange dual as used in~\cite{efficientsparsecoding_nips}. Throughout this paper, parameter $\lambda$ is set 0.1. In our experiment, the dictionaries learned contain 1500 visual words for Brodatz and KTH-TIPS dataset, and $3000$ visual words for UMD dataset.

\textbf{Partition strategy and overlapping patterns.}
For the experiments, we partition all the texture images into 4 levels ($2\times2, 3\times3, 4\times 4, 5\times 5$) over Brodatz dataset, 3 levels ($ 6\times6, 7\times7, 8\times 8 $) for KTH-TIPS dataset, and 4 levels ($3\times3, 4\times 4, 5\times 5, 6\times6$) for UMD texture database. Furthermore, over each partition level, we admit various overlapping patterns. Actually, we empirically find the partition strategy of each three datasets produces satisfactory results.

\textbf{LC-CRC framework for classification. }
In our proposed LC-CRC classification framework, there is a parameter $\lambda$ (different from the one of dictionary learning in Equation~\ref{equa:DL}) to make the solution of the least square problem Equation~\ref{equa:CRC_RLS} stable. Through empirical observations, we find that the experimental results are not sensitive to the choice of $\lambda$ if a small value is assigned which is less than $0.01$, and thus we set $\lambda$ as $0.001$ through out our work. Moreover, parameter $K$ of KNN algorithm ought to be specified, and we set $K=100$ when the number of training samples per class is small, \eg only 1 or 2 samples of each class are available for training, and $K=300$ when more training samples per class are available.

%------------------------------------------------------------------------
\subsection{Brodatz dataset}
\label{ssec:Brodatz}

\begin{figure*}[t]
\begin{center}
   \includegraphics[width=1\linewidth]{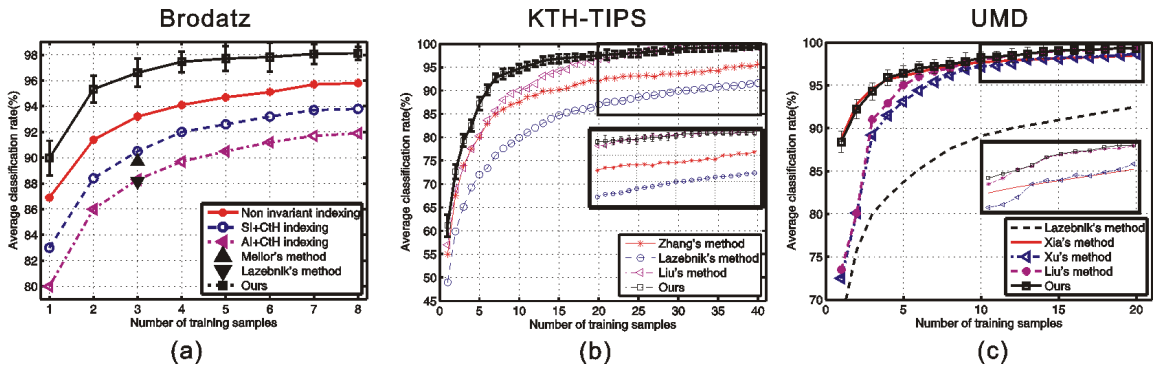}
\end{center}
   \caption{Classification rate vs. number of training samples on the three datasets.}
\label{fig:Three_lineChart}
\end{figure*}

Figure~\ref{fig:Three_lineChart}~(a) shows the classification results obtained on the Brodatz database. Following Lazebnik \etal~\cite{UIUC_texture_PAMI2005} and Mellor \etal~\cite{LocallyInvariant_Texture_PAMI2008}, classification rates are estimated by averaging the results on randomly selected training sets, and 10 trials are performed in our experiments.

SI+CtH indexing and AI+CtH indexing are two shape-based invariant texture features from the work of Xia \etal~\cite{shape_Texture_IJCV2010}. Here SI proposed by Xia \etal is a kind of feature that is invariant to (local) similarity transforms, and AI means (locally) affine invariant features. Both of them are made of several histograms, such as scale ratio histogram, elongation histogram and compactness histogram. Moreover, CtH is contrast histogram computed by scanning all pixels of a local adaptive neighborhood, which is robust to geometrical distortions of the textures~\cite{shape_Texture_IJCV2010}. Due to that the samples in this dataset are created by cutting each texture  of the Brodatz database into pieces, as a consequence, the resulting dataset lacks of viewpoint and scale changes. For this reason, Xia \etal also adopt a well chosen non-invariant indexing scheme (Non-invariant indexing in Figure~\ref{fig:Three_lineChart}~(a)) and it shows better classification result. Despite multiple histograms in~\cite{shape_Texture_IJCV2010}, our framework only employs one kind of feature descriptor (SIFT), and it achieves state-of-the-art performance.
Note that when $3$ samples per class are used for training, our approach achieves $96.61\%$. This outcome is higher than $95.9\%$ achieved by the method of Zhang \etal, based on the method of Lazebnik \etal~\cite{UIUC_texture_PAMI2005} by employing three types of descriptors (SPIN, RIFT and SIFT)~\cite{LocalFeatures_IJCV2007}, and is comparable with $97.16\%$ (the highest classification rate on Brodatz dataset to the best of our knowledge) attained by the method of Liu \etal by using sorted random projections plus several kernel SVMs~\cite{SortedRandomProjection_Texture_ICCV2011}. However, it is worth noticing that when only one images of each class is used for training, our approach achieves $90\%$ accuracy, which is significantly higher than the other methods. This verifies that our approach indeed can extract large amount reliable features of each type of textures, even when only a few sample images are available for training.

\subsection{KTH-TIPS texture database}
\label{ssec:KTHTIPS}

Following Zhang \etal~\cite{LocalFeatures_IJCV2007}, we vary the number of training images and record classification accuracy, as Figure~\ref{fig:Three_lineChart}~(b) shows. Note that all images are converted to grey scale in our approach and no use of color information is made whatsoever. Three methods are used for comparison, and the results of these methods are taken directly from the original publications or quoted from the recent comparative study of Zhang \etal~\cite{LocalFeatures_IJCV2007}. In~\cite{UIUC_texture_PAMI2005}, Lazebnik \etal first characterize the texture using Harris-affine corners and Laplacian-affine blobs with two descriptors (SPIN and RIFT), and employ nearest neighbor classifier. Their method achieves $91.3\%$ accuracy when 41 samples of each class are used for training. Under the same configuration, the method of Zhang etal, introduced in Subsection~\ref{ssec:Brodatz}, achieves $96.1\%$, and the approach of Liu \etal achieves $99.29\%$ (the highest classification rate on KTH-TIPS dataset to the best of our knowledge) in~\cite{SortedRandomProjection_Texture_ICCV2011}. And our method achieves $99.32\%$ under the condition that 41 samples per material are used for training, which exceeds the best one ($99.29\%$).

\begin{figure}[t]
\begin{center}
   \includegraphics[width=0.70\linewidth]{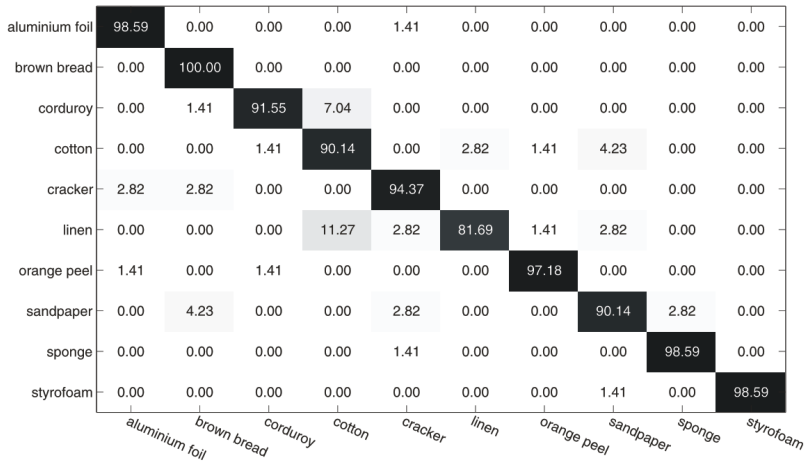}
\end{center}
   \caption{Performance on KTH-TIPS texture database, confusion matrix for classification of 10 different textures. $10$ images per class are randomly selected for training, and the rest for testing. The number at row R and column C is the proportion of R class which is classified as C class. For example, $9.86\%$ of the linen images are misclassified as cotton class. The average accuracy is $94.23\%$.}
\label{fig:KTHTIPS_confusionMatrix}
\end{figure}

\begin{figure}[t]
\begin{center}
   \includegraphics[width=0.70\linewidth]{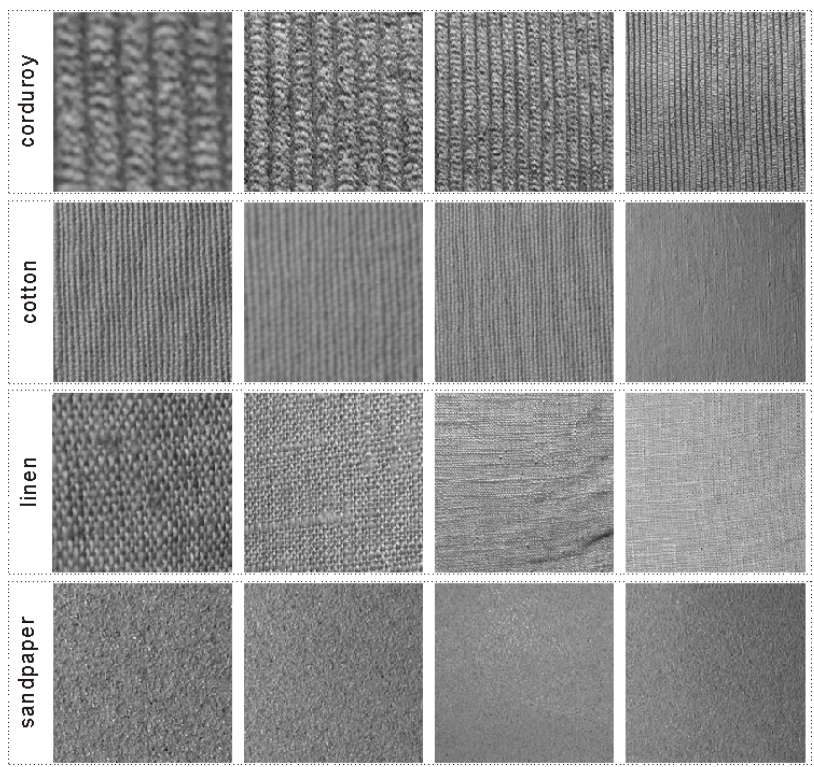}
\end{center}
   \caption{Similar texture pattern on KTH-TIPS database. Four different texture types are displayed here, it is easy to see that some images of the four textures are very similar with various scales. This phenomenon largely leads to misclassification on this dataset.}
\label{fig:similarPattern_KTHTIPS}
\end{figure}

It is worth noting that our approach achieves $(94.1\pm 0.92)\%$ when only 10 images of each class are randomly selected for training, which is significantly higher than the others. Figure~\ref{fig:KTHTIPS_confusionMatrix} displays one confusion matrix under this condition. From the confusion matrix, we can see the misclassifications mainly concentrate on four materials: corduroy, cotton, linen and sandpaper. Figure~\ref{fig:similarPattern_KTHTIPS} shows some samples of the four types of materials, and it can be easily seen that under different scale of different materials, they are very similar and this phenomenon results misclassification within these material types.

%------------------------------------------------------------------------
\subsection{UMD texture database}
\label{ssec:UMD}

The UMD texture database contains images of larger arbitrary rotation, larger scale variation and more significant viewpoint than the previous two datasets. Therefore, it is more challenging for classification.

Figure~\ref{fig:Three_lineChart}~(c) shows the classification rate vs. the number of training samples on UMD dataset. Xia's method denotes the SI+CtH indexing method as described in Subsection~\ref{ssec:Brodatz} in conjunction with geodesic distance, which considers textures as points lying on some intrinsic manifold and yields clear improvement in their method. Xu's method is based on a combination of wavelet transform and multifractal analysis. Liu's method is introduced previously in Subsection~\ref{ssec:Brodatz}.
We can see when only a few samples of each class are available for training, our method is comparable to Xia's method, which achieves the best performance on this database under small amount of training samples. While the number of training images of each class is increasing, Liu's method obtains better results. Still, Under this condition, our method achieves comparable outcome with Liu's method. When 20 sample images per class are randomly selected for training,  $98.6\%$ classification accuracy is achieved by Xu's method and $99.30\%$ by Liu's method (the best one ever reported on this database). And our method achieves $(99.32 \pm 0.35)\%$ classification accuracy, which is slightly better than Liu's result.

From this experiment on UMD dataset, we can see our proposed texture classification approach can extract reliable texture features while only a few training sample images are available and leads to significantly better results. While the number of training sample grows, our method can still achieve state-of-the-art performance compared with other methods. In the consideration of the complex texture sample images from UMD dataset, it is easy to see that our method achieves invariance to local rotation variation, scale changes, translation, changes of illumination directions and significant viewpoint.

%------------------------------------------------------------------------
\section{Conclusion and Future Work}
\label{sec:conclusion}

In this paper, focusing on texture classification task, we introduce a novel and highly effective scheme for robust texture classification, which is invariant to scale differences, translation, significant viewpoint changes and local rotation. Inspired by SPM framework, we first develop a multi-level descriptor to describe local texture features, allowing different level of partitions and various overlapping patterns within each level of partition. From experiments, we see this flexible descriptor can better capture the local features of each kind of texture, and even when only a few samples of each class are available for training, our method still achieves very high accuracy. Accordingly, we propose an efficient classification mechanism, which is based on collaborative representation with locality constraint, called LC-CRC. It first search relatively a few neighbors from the feature pond by KNN algorithm, and then use them to represent the target through solving a simple least square fitting problem with $\ell_2$-norm regularization. To evaluate our texture classification framework, we conduct several experiments on three well-known texture datasets and the outcome is very competitive and even outperforms several state-of-the-art methods.

Actually, LC-CRC classification framework treat the feature pond as another dictionary, which is used to represent the pooled feature codes of testing images. This spirit of hierarchical sparse coding has been already explored by Yu \etal in~\cite{hierarchicalSC_CVPR2011} for object recognition, but there remains interesting extensions and confirmations, and one of our future work is to provide some insights of multi-layer dictionary learning for image classification.
Moreover, our work provides a new application of SPM, and we expect some other applications based on SPM and its variants.

\subsection*{Acknowledgements}
This work is supported by by 973 Program (Project No.2010CB327905) and Natural Science Foundations of China (No.61071218).

\bibliography{bib}

\begin{thebibliography}{10}

\bibitem{Xu_IJCV2009}
Y.~Xu, H.~Ji, and C.~Fermuller, ``Viewpoint invariant texture description using
  fractal analysis,'' {\em IJCV}, 2009.

\bibitem{WaveletPyramid_CVPR2010}
Y.~Xu, X.~Yang, H.~Ling, and H.~Ji, ``A new texture descriptor using
  multifractal analysis in multi-orientation wavelet pyramid,'' {\em CVPR},
  2010.

\bibitem{SortedRandomProjection_Texture_ICCV2011}
L.~Liu, P.~Fieguth, G.~Kuang, and H.~Zha, ``Sorted random projections for
  robust texture classification,'' {\em ICCV}, 2011.

\bibitem{Lazebnik_PAMI2005}
S.~Lazebnik, C.~Schmid, and J.~Ponce, ``A sparse texture representation using
  local affine regions,'' {\em PAMI}, 2005.

\bibitem{LocalFeatures_IJCV2007}
J.~Zhang, M.~Marszalek, S.~Lazebnik, and C.~Schmid, ``Local features and
  kernels for classification of texture and object categories: A comprehensive
  study,'' {\em IJCV}, 2007.

\bibitem{basicFeatureforTextureClassification_IJCV2010}
M.~Crosier and L.~D. Griffin, ``use basic image features for texture
  classification,'' {\em IJCV}, 2010.

\bibitem{Lazebnik_CVPR06}
S.~Lazebnik, C.~Schmid, and J.~Ponce, ``Beyond bags of features: Spatial
  pyramid matching for recognition natural scene categories,'' {\em CVPR},
  2006.

\bibitem{Lowe_IJCV2004}
D.~G. Lowe, ``Distinctive image features from scale-invariant keypoints,'' {\em
  IJCV}, 2004.

\bibitem{HOG_cvpr2005}
N.~Dalal and B.~Triggs, ``Histograms of oriented gradients for human
  detection,'' {\em CVPR}, 2005.

\bibitem{SPM_Yang_cvpr2009}
J.~Yang, K.~Yu, Y.~Gong, and T.~Huang, ``Linear spatial pyramid matching using
  sparse coding for image classification,'' {\em CVPR}, 2009.

\bibitem{Boureau_locals_ICCV2011}
Y.-L. Boureau, N.~L. Roux, F.~Bach, J.~Ponce, and Y.~LeCun, ``Ask the locals:
  multi-way local pooling for image recognition,'' {\em ICCV}, 2011.

\bibitem{discriminativeSpatialPyramid_cvpr2011}
T.~Harada, Y.~Ushiku, Y.~Yamashita, and Y.~Kuniyoshi, ``Discriminative spatial
  pyramid,'' {\em CVPR}, 2011.

\bibitem{SRC_pami2009}
J.~Wright, A.~Y. Yang, A.~Ganesh, S.~S. Sastry, and Y.~Ma, ``Robust face
  recognition via sparse representation,'' {\em PAMI}, 2008.

\bibitem{Grauman_ICCV05}
K.~Grauman and T.~Darrell, ``The pyramid match kernel: Discriminative
  classification with sets of image feature,'' {\em ICCV}, 2005.

\bibitem{Boureau_MidLevelFeature_CVPR2010}
Y.-L. Boureau, F.~Bach, Y.~LeCun, and J.~Ponce, ``Learning mid-level features
  for recognition,'' {\em CVPR}, 2010.

\bibitem{WhatIsTexture}
Y.~Meyer, ``Workshop: An interdisciplinary approach to textures and natural
  images processing,'' {\em Institut Henri Poincar$\acute e$, Paris}, 2007.

\bibitem{FaceRec_CSscheme_CVPR2011}
Q.~Shi, A.~Eriksson, A.~van~den Hengel, and C.~Shen, ``Is face recognition
  really a compressive sensing problem?,'' {\em CVPR}, 2011.

\bibitem{FaceRec_SRorCollaborate_CVPR2011}
L.~Zhang, M.~Yang, and X.~Feng, ``Sparse representation or collaborative
  representation: Which helps face recognition?,'' {\em ICCV}, 2011.

\bibitem{NonlinearLLC_NIPS2009}
K.~Yu, T.~Zhang, and Y.~Gong, ``Nonlinear learning using local coordinate
  coding,'' {\em NIPS}, 2008.

\bibitem{LLC_Wang_cvpr2010}
J.~Wang, J.~Yang, K.~Yu, F.~Lv, T.~Huang, and Y.~Gong, ``Learning
  locality-constrained linear coding for image classification,'' {\em CVPR},
  2010.

\bibitem{Boureau_ICML2010}
Y.-L. Boureau, J.~Ponce, and Y.~Lecun, ``A theoretical analysis of feature
  pooling in visual recognition,'' {\em ICML}, 2010.

\bibitem{SupTranInvariantSC_CVPR2010}
J.~Yang, K.~Yu, and T.~Huang, ``Supervised translation-invariant sparse
  coding,'' {\em CVPR}, 2010.

\bibitem{Brodatz1966}
P.~Brodatz, ``Textures: A photographic album for artists and designers,'' {\em
  New York: Dover}, 1966.

\bibitem{KTH_TIPS_ECCV2004}
E.~Hayman, B.~Caputo, M.~Fritz, and J.-O. Eklundh, ``On the significance of
  real-world conditions for material classification,'' {\em ECCV}, 2004.

\bibitem{efficientsparsecoding_nips}
H.~Lee, A.~Battle, R.~Raina, and A.~Y. Ng, ``Efficient sparse coding
  algorithms,'' {\em NIPS}, 2007.

\bibitem{UIUC_texture_PAMI2005}
J.~P. Svetlana~Lazebnik, Cordelia~Schmid, ``A sparse texture representation
  using local affine regions,'' {\em PAMI}, 2005.

\bibitem{LocallyInvariant_Texture_PAMI2008}
M.~Mellor, B.-W. Hong, and M.~Brady, ``Locally rotation, contrast, and scale
  invariant descriptors for texture analysis,'' {\em TPAMI}, 2008.

\bibitem{shape_Texture_IJCV2010}
G.-S. Xia, J.~Delon, and Y.~Gousseau, ``Shape-based invariant texture
  indexing,'' {\em IJCV}, 2010.

\bibitem{hierarchicalSC_CVPR2011}
K.~Yu, Y.~Lin, and J.~Lafferty, ``Learning image representations from the pixel
  level via hierachical sparse coding,'' {\em CVPR}, 2011.

\end{thebibliography}
\bibliographystyle{hieeetr} % ieee

\end{document}